\documentclass[runningheads]{llncs}
\usepackage[T1]{fontenc}
\usepackage{marvosym}
\usepackage{graphicx}
\usepackage{xcolor}
\usepackage{amsmath}
\usepackage{amssymb}
\usepackage{pifont}
\usepackage{orcidlink}
\hypersetup{hidelinks,pdftitle={CasCVS-Net: A Staged Multi-Task Cascade for Critical View of Safety Assessment},pdfauthor={Bock-Zien Toh, Yuanchuan Ren, Tay Aw Yu, Ng Khee Ong, Zhehua Mao, Sophia Bano}}
\makeatletter
\patchcmd{\@maketitle}{\newpage}{\newpage\vspace*{-61.25pt}}{}{
  \PackageError{cascvs}{Unable to adjust title spacing}{}
}
\makeatother
\begin{document}
\flushbottom
\title{CasCVS-Net: A Staged Multi-Task Cascade for Critical View of Safety Assessment}
\titlerunning{CasCVS-Net}
\author{Bock-Zien Toh\inst{1}$^{\dagger}$ \and
Yuanchuan Ren\inst{1}$^{\dagger}$ \and
Tay Aw Yu\inst{1} \and
Ng Khee Ong\inst{1} \and Zhehua Mao\inst{1,2}\orcidlink{0000-0001-9225-6318}\Letter \and Sophia Bano\inst{1,2}\orcidlink{0000-0003-1329-4565}}
\authorrunning{Toh, Ren et al.}
\institute{
Department of Computer Science, University College London, London, UK \and
UCL Hawkes Institute, University College London, UK\\
$^{\dagger}$ These authors contributed equally to this work.\\
\Letter\ Corresponding author: \email{z.mao@ucl.ac.uk}
}
\maketitle
\begin{abstract}
Automated assessment of the Critical View of Safety (CVS) in laparoscopic cholecystectomy requires both recognition of the three CVS criteria and anatomical grounding in small, rare, and often occluded hepatocystic structures. Learning-based methods differ in the anatomical information they use, from image-level classification to detection, segmentation, or graph-based reasoning, yet grounding the safety-critical anatomy remains the main bottleneck. We propose CasCVS-Net, a staged multi-task cascade that jointly performs object detection, semantic segmentation, and CVS assessment, trained on the Endoscapes dataset. The model couples the tasks through predicted anatomy: predicted boxes guide segmentation, and predicted masks provide region-level features for CVS classification, so CVS assessment at inference uses only model predictions rather than ground-truth annotations. To reduce optimisation instability in this coupled setting, training progresses from detection to detection--segmentation and then to the full three-task cascade, followed by task-wise fine-tuning. Evaluation on the public unseen test set shows that CasCVS-Net improves over matched single-task baselines on all three tasks, achieving 32.0 detection mAP, 46.8 semantic mIoU, 15.3 rare-anatomy mIoU, and 67.2 CVS mAP. It outperforms the state-of-the-art LG-CVS and SV2LSTG by 6.3\% and 4.5\% relative CVS mAP, respectively, corresponding to 4.0 and 2.9 mAP points. These results show that staged task coupling through predicted boxes and masks improves anatomical grounding for CVS assessment, particularly for rare hepatocystic structures.

\keywords{Critical View of Safety \and Staged multi-task learning \and Multi-task learning \and Surgical scene understanding}

\end{abstract}
\section{Introduction}
\label{sec:intro}

Bile duct injury is among the most serious complications of laparoscopic cholecystectomy, occurring in roughly 0.3--0.7\% of cases and carrying high morbidity and cost~\cite{brunt2020safe}. Many injuries arise from misidentification of biliary anatomy rather than technical execution alone~\cite{way2003causes}. The Critical View of Safety (CVS), introduced by Strasberg et al.~\cite{strasberg1995analysis}, aims to prevent such errors by requiring three conditions before division of the cystic duct or artery: only the cystic duct and cystic artery enter the gallbladder (C1), the hepatocystic triangle is cleared (C2), and the lower gallbladder is separated from the cystic plate (C3). Although reliable CVS achievement is associated with safer surgery~\cite{manatakis2023cvs}, CVS documentation remains inconsistent~\cite{mascagni2020formalizing}, motivating automated assessment that is both accurate and anatomically grounded.

CVS assessment is commonly formulated as multi-label prediction of the three CVS criteria, and learning-based methods differ mainly in how they use anatomical supervision or representation. Segmentation-based pipelines segment the hepatocystic anatomy before classification~\cite{mascagni2022deepcvs}; detection- and graph-based methods localise anatomy with bounding boxes and reason over the detected structures~\cite{murali2023lgcvs,murali2023sv2lstg}; other methods detect landmarks or safety zones for intraoperative guidance~\cite{madani2022gonogonet,tokuyasu2021landmarks}; and more recent transformer- or representation-learning classifiers operate directly at the image level~\cite{nowak2025swincvs,baby2025cvsadaptnet}. Despite this progress, reliable anatomical grounding remains challenging, especially for the rare and small safety-critical structures such as the cystic plate and hepatocystic triangle, which are difficult to localise and segment. This motivates a detection-guided cascade, in which detection provides coarse localisation cues for segmentation, and segmentation in turn provides an anatomy-aware representation for CVS assessment.

We propose \emph{CasCVS-Net}, a staged multi-task cascade linking object detection, semantic segmentation, and CVS prediction in a shared ResNet-50--FPN network with task-specific residual adapters. The framework is trained progressively: a detection head is trained first, and its predicted boxes are converted into class-aware spatial priors that guide training of the segmentation head; the predicted semantic masks then define the anatomical regions utilized by a mask-pooled CVS head. Because each head is trained on the same predicted boxes and masks it will receive at test time, the cascade needs no ground-truth boxes or masks at inference.

Our key contributions are: (i) a detection-guided segmentation cascade that uses predicted boxes as coarse priors for rare safety-critical anatomy; (ii) a mask-pooled CVS head that extracts region-level features from predicted semantic masks, keeping the training and inference pathways consistent; and (iii) an Endoscapes evaluation that reports detection and segmentation for the rare safety-critical anatomy separately, with ablations isolating staged training, box-to-mask guidance, and mask-to-CVS coupling.

\section{Method}
\label{sec:method}

Given a laparoscopic frame, the model predicts object detections, a semantic segmentation map, and CVS criterion scores. Detection uses six foreground classes: cystic plate, hepatocystic triangle, cystic artery, cystic duct, gallbladder, and surgical instrument. Segmentation predicts the same six classes plus background. The CVS branch predicts the three criteria C1--C3.

\begin{figure}[!t]
\centering
\includegraphics[width=0.75\textwidth]{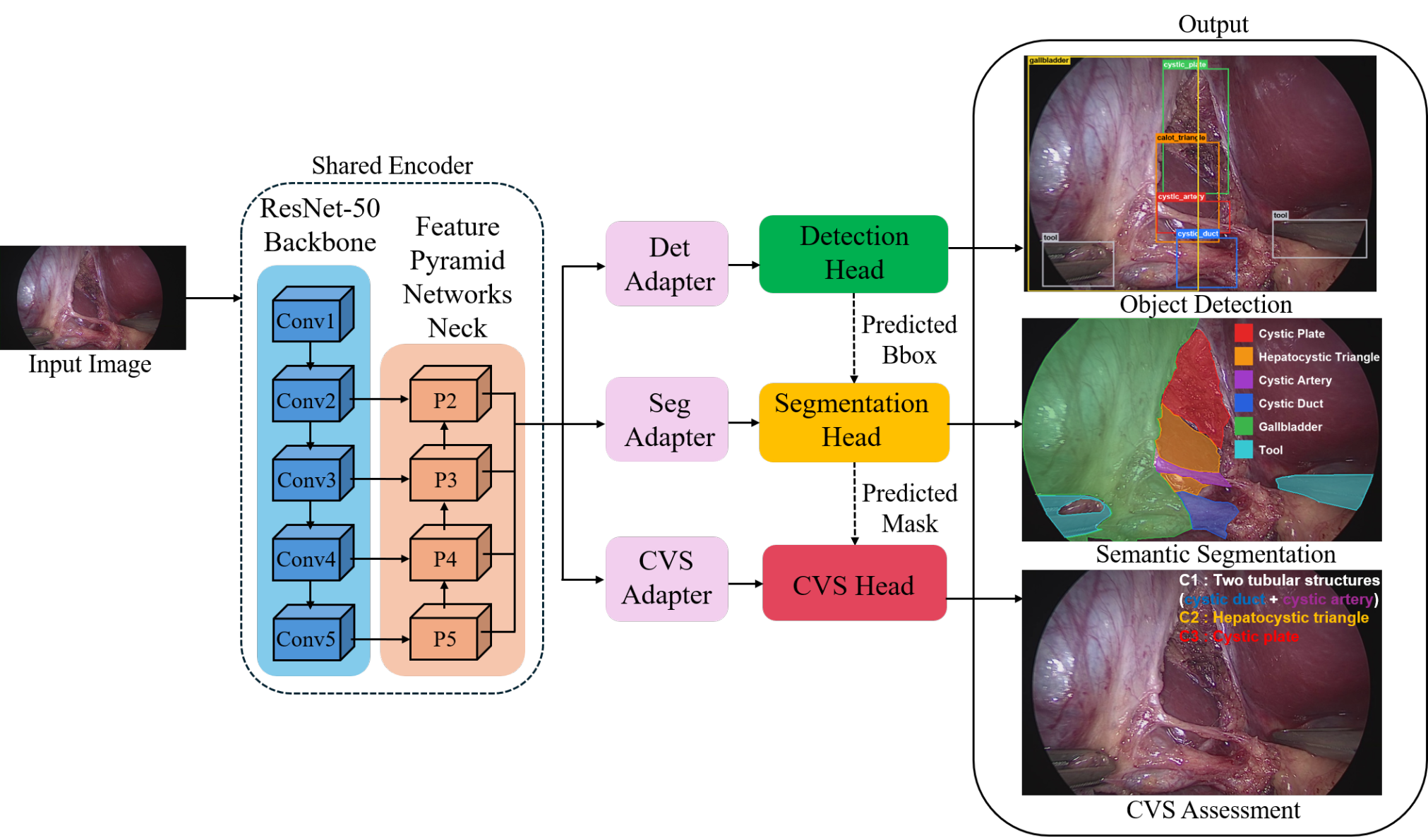}
\caption{Overview of the proposed staged cascade. A shared ResNet-50--FPN encoder produces multi-scale features, which are refined by task-specific adapters before the detection, segmentation, and CVS heads. At inference, predicted boxes guide segmentation and predicted masks define regions for CVS feature pooling.}
\label{fig:model_architecture}
\end{figure}

The model is a cascaded multi-task predictor with a shared ResNet-50--FPN encoder~\cite{he2016resnet,lin2017fpn} and task-specific heads (Fig.~\ref{fig:model_architecture}), producing multi-scale pyramid features. Before each head, a lightweight residual adapter~\cite{rebuffi2017adapters} refines every FPN level: a $1\times1$ then $3\times3$ convolution with a ReLU between, the final layer zero-initialised and added back to the level's features, so each adapter starts as the identity and learns only a task-specific residual. The three task adapters share this design but keep independent weights.

The detection branch is a Faster R-CNN head~\cite{ren2015faster}: a region proposal network generates candidate boxes on every pyramid level, and a region-of-interest head classifies and refines them into the six anatomy and instrument classes. The segmentation branch is a DeepLabV3+ decoder~\cite{chen2018deeplab}, and the CVS branch is the mask-pooled head of Section~\ref{sec:mask_pooled_cvs}, each on its adapted features. At inference these heads form a cascade (Sections~\ref{sec:box_to_mask} and~\ref{sec:mask_pooled_cvs}): the model maps a frame to boxes, then masks, then CVS scores using only its own predictions.

\subsection{Box-to-Mask Coupling}
\label{sec:box_to_mask}

The first cascade coupling converts detector outputs into a spatial prior for segmentation. From the detections $\mathcal{D}=\{(b_i,c_i,s_i)\}_{i=1}^{N}$, with box $b_i$, predicted class $c_i$, and confidence $s_i$, we retain those with $s_i\geq0.1$, enlarge each box by $10\%$, and clip it to the image boundaries. The threshold is deliberately low because the rare safety-critical structures are detected with low confidence: a stricter cut-off would drop them from the prior. We instead keep them and let the confidence weighting and the learned zero-initialised projector discount the resulting false positives, which recovers more rare anatomy. The enlarged boxes are rasterised into a seven-channel prior $B$ whose background channel stays zero and whose six foreground channels are confidence-weighted:
\begin{equation}
B_c(p)=
\max\Bigl(\{s_i \mid c_i=c,\; p\in b'_i\}\cup\{0\}\Bigr),
\quad c\in\{1,\ldots,6\},
\label{eq:prior_max}
\end{equation}
where $b'_i$ is the enlarged box and $p$ indexes pixels, so same-class overlaps take the maximum confidence while different classes occupy separate channels. In the joint pretraining stage the values are binary, so this maximum reduces to a union of the overlapping boxes. The prior is resized, passed through a convolutional projector initialised to output zero, and added as a residual to the decoder feature of the DeepLabV3+ segmentation head, which produces per-pixel logits over the seven classes; the prior begins as a no-op and is learned as a correction. Predicted detections are detached before prior construction, so segmentation loss does not backpropagate through the detector.

\subsection{Mask-Pooled CVS Head}
\label{sec:mask_pooled_cvs}

\begin{figure}[!t]
    \centering
    \includegraphics[width=0.8\textwidth]{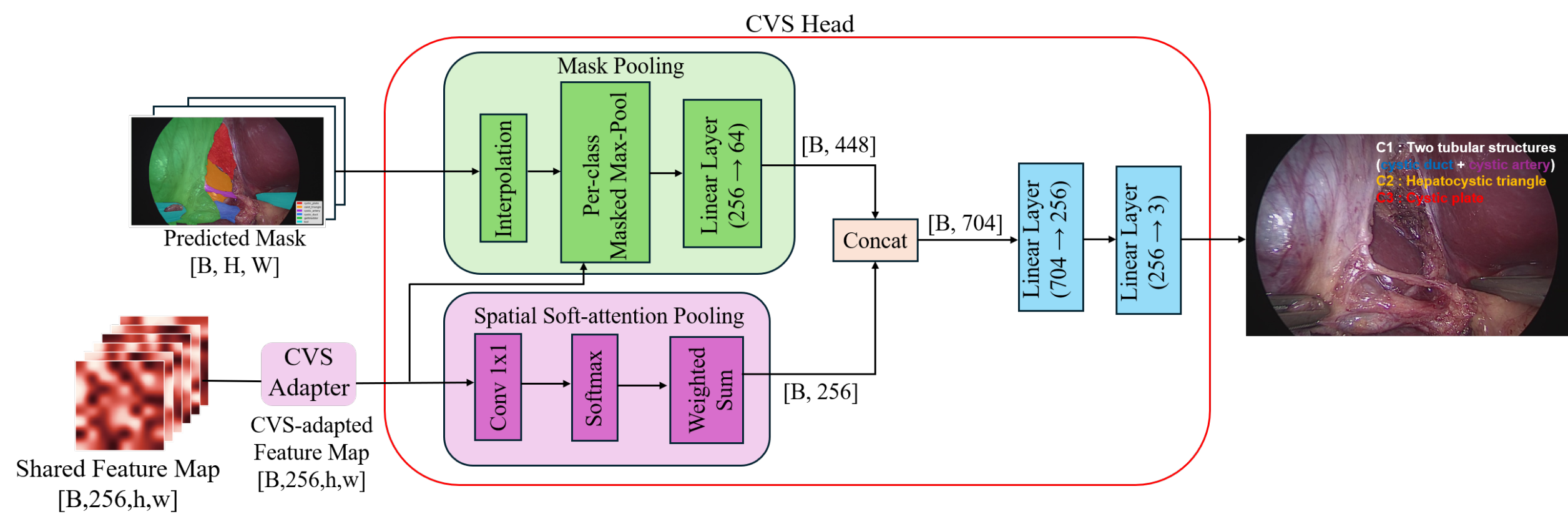}
    \caption{Mask-pooled CVS head. The predicted semantic map is converted into hard class regions and downsampled to the CVS feature grid. Class-wise mask pooling extracts anatomical embeddings, while a spatial attention branch summarises global context. The concatenated representation is passed through fully connected layers to predict the three CVS criteria. Here $B$ denotes the batch size; the mask branch stacks one $64$-dimensional embedding per class ($7\times64=448$).}
    \label{fig:cvs_head}
\end{figure}

The second coupling lets the CVS head utilize the predicted anatomy rather than only global image features.

The segmentation logits are reduced to a hard label map by a per-pixel arg-max over the seven classes and downsampled to the CVS feature grid. Each class $c$ then defines a region $\mathcal{R}_c$ of grid positions $q$ labelled $c$ over the CVS-adapted feature map $F_{\mathrm{cvs}}$, and its embedding is the per-channel maximum $e_c=\max_{q\in\mathcal{R}_c}F_{\mathrm{cvs}}(q)$ when $\mathcal{R}_c\neq\emptyset$ and $e_c=\mathbf{0}$ otherwise. Stacking the per-class embeddings, one slot per semantic class including background, yields the mask feature.

In parallel, a soft spatial-attention branch pools the same features into a global summary: a $1\times1$ convolution reduces the feature map to a single-channel score, and a softmax over the grid gives the attention weights that sum the features. The per-class mask embeddings and this summary are concatenated and passed through fully connected layers to produce the three CVS logits.

Because the regions come from a hard $\arg\max$ semantic map, CVS loss does not backpropagate through mask membership into the segmentation decoder. The coupling therefore uses segmentation as a structured anatomical representation, while CVS supervision updates the CVS branch and any shared features trainable in the current stage.

\subsection{Staged Training Strategy}
\label{sec:training}

Training the three tasks jointly from scratch is unstable: detection, dense segmentation, and image-level CVS impose conflicting demands on the shared encoder, and a weak detector provides unreliable inputs to the downstream heads. We therefore build the cascade in four stages. \textbf{(i)~Detection:} we train the Faster R-CNN branch alone to establish the localisation and multi-scale features the cascade depends on. \textbf{(ii)~Detection--segmentation:} we add the segmentation branch and train it together with detection, warm-starting its head from a pre-trained single-task segmentation model; the FPN, both heads, the adapters, and the late ResNet stages update while the early stages stay frozen. \textbf{(iii)~Full cascade:} we add the CVS head and train all three tasks together with both couplings active, so each head learns from the predicted boxes and masks it will see at inference (this stage uses the binary box prior of Section~\ref{sec:box_to_mask}); PCGrad~\cite{yu2020pcgrad} removes the conflicting component of each task gradient on the shared backbone and FPN before the update. \textbf{(iv)~Fine-tuning:} finally, we fine-tune the heads one at a time in cascade order (detection, segmentation, CVS). Each pass trains only that head and its adapter with the rest frozen, recalibrating it on the now-fixed upstream predictions; the segmentation pass switches the box prior from its binary form to the confidence-weighted form of Eq.~\ref{eq:prior_max}. This preserves the shared representation.

\subsection{Training Objective}
\label{sec:objective}

Each task head is trained with its own loss; because annotation availability differs by task, each sample contributes only to the losses for which its label is present. Detection uses the Faster R-CNN multi-task loss~\cite{ren2015faster}, the sum of the region-proposal-network and detection-head classification (cross-entropy) and box-regression (smooth-$L_1$) terms, $\mathcal{L}_{\mathrm{det}}=\mathcal{L}^{\mathrm{cls}}_{\mathrm{rpn}}+\mathcal{L}^{\mathrm{reg}}_{\mathrm{rpn}}+\mathcal{L}^{\mathrm{cls}}_{\mathrm{roi}}+\mathcal{L}^{\mathrm{reg}}_{\mathrm{roi}}$. Segmentation uses a pixel-wise cross-entropy over the $K=7$ classes, averaged over the valid pixels $\Omega$, $\mathcal{L}_{\mathrm{seg}}=-|\Omega|^{-1}\sum_{p\in\Omega}\log\mathrm{softmax}_{y_p}(s_p)$, where $s_p\in\mathbb{R}^{K}$ are the class logits at pixel $p$ and $y_p$ its ground-truth label. CVS uses a class-balanced binary cross-entropy over the three criteria, $\mathcal{L}_{\mathrm{cvs}}=-\frac{1}{3}\sum_{c=1}^{3}[w_c t_c\log\sigma(z_c)+(1-t_c)\log(1-\sigma(z_c))]$, with criterion logits $z_c$, soft consensus targets $t_c\in[0,1]$, the sigmoid $\sigma$, and inverse-frequency positive weights $w_1=5.40$, $w_2=7.92$, and $w_3=4.59$. The joint cascade stage minimises the fixed weighted sum $\mathcal{L}=2.0\mathcal{L}_{\mathrm{det}}+1.5\mathcal{L}_{\mathrm{seg}}+0.5\mathcal{L}_{\mathrm{cvs}}$, whose weights bring the three raw-loss magnitudes onto a comparable scale (the CVS term is largest, segmentation smallest); each fine-tuning stage instead optimises only its active head's loss.

\section{Experiments}
\label{sec:experiments}

\noindent\textbf{Dataset.} We evaluate on Endoscapes2023, a publicly available laparoscopic cholecystectomy dataset with aligned annotations for object detection, semantic segmentation, and CVS assessment~\cite{murali2023endoscapes,mascagni2025endoscapes}. The dataset contains 201 videos with an official video-level split of 120 training, 41 validation, and 40 test videos. CVS labels, taken as the consensus of three expert annotators, are available for 11{,}090 frames, bounding boxes for 1{,}933 frames, and public semantic masks for the Endoscapes-Seg50 subset. Detection and segmentation use the same six foreground classes; segmentation additionally includes background. The safety-critical anatomy is both rare and hard to localise: on the test split, the cystic plate and hepatocystic triangle appear in only 41.7\% and 39.7\% of frames, and their detection mAP is 8--19, compared with over 60 for the gallbladder and surgical instrument~\cite{murali2023endoscapes}.

\noindent\textbf{Metrics.} Detection is evaluated with COCO-style bounding-box\hfill\break mAP@[0.5:0.95]. Segmentation is evaluated with semantic mean intersection-over-union (mIoU) over the seven classes. CVS is evaluated by average precision per criterion and mean AP over C1--C3; balanced accuracy uses thresholds calibrated on the validation split and applied to the held-out test split. Because these structures are the safety-critical bottleneck, we additionally report rare detection mAP and rare segmentation mIoU over the cystic plate and hepatocystic triangle. Since our segmentation metric is semantic mIoU rather than the official Endoscapes instance-segmentation AP, segmentation comparisons are restricted to our own framework-matched baselines.

\noindent\textbf{Baselines.} The primary controls are framework-matched single-task learning (STL) baselines for detection, segmentation, and CVS, each using the same ResNet-50--FPN encoder and corresponding task head as the proposed model. For CVS, we also compare against published LG-CVS~\cite{murali2023lgcvs} and SV2LSTG~\cite{murali2023sv2lstg} benchmark results under the same public annotation setting. We do not compare against results using the private Endoscapes-Seg201 annotations, which are not publicly available.

\noindent\textbf{Implementation details.} We implement all models in PyTorch. A shared preprocessing pipeline applies the same geometric transforms to images, boxes, and masks. During training, the input scale is sampled from $854\times480$, $1024\times576$, and $1280\times720$ at a fixed aspect ratio, and evaluation is bounded by $1280\times720$. Semantic masks are resized by nearest neighbour, images are normalised with ImageNet statistics, and augmentation is horizontal flip, translation, colour jitter, and cutout.

The full cascade trains with SGD (learning rate $0.0015$, momentum $0.9$, weight decay $5\times10^{-4}$, batch size 2) for up to 16 epochs with validation-based checkpointing, freezing all but the last two ResNet stages and disabling mixed precision; the detection, segmentation, and CVS fine-tuning stages use learning rate $0.0008$ for up to 12, 12, and 6 epochs. At inference, Faster R-CNN applies class-wise NMS at $0.5$, score threshold $0.05$, and at most 100 detections per image; the segmentation prior follows Section~\ref{sec:box_to_mask}. Runtime and memory profiling are left for future deployment-oriented evaluation.

\section{Results and Discussion}
\label{sec:results}

\begin{figure}[!t]
    \centering
    \includegraphics[width=0.8\textwidth]{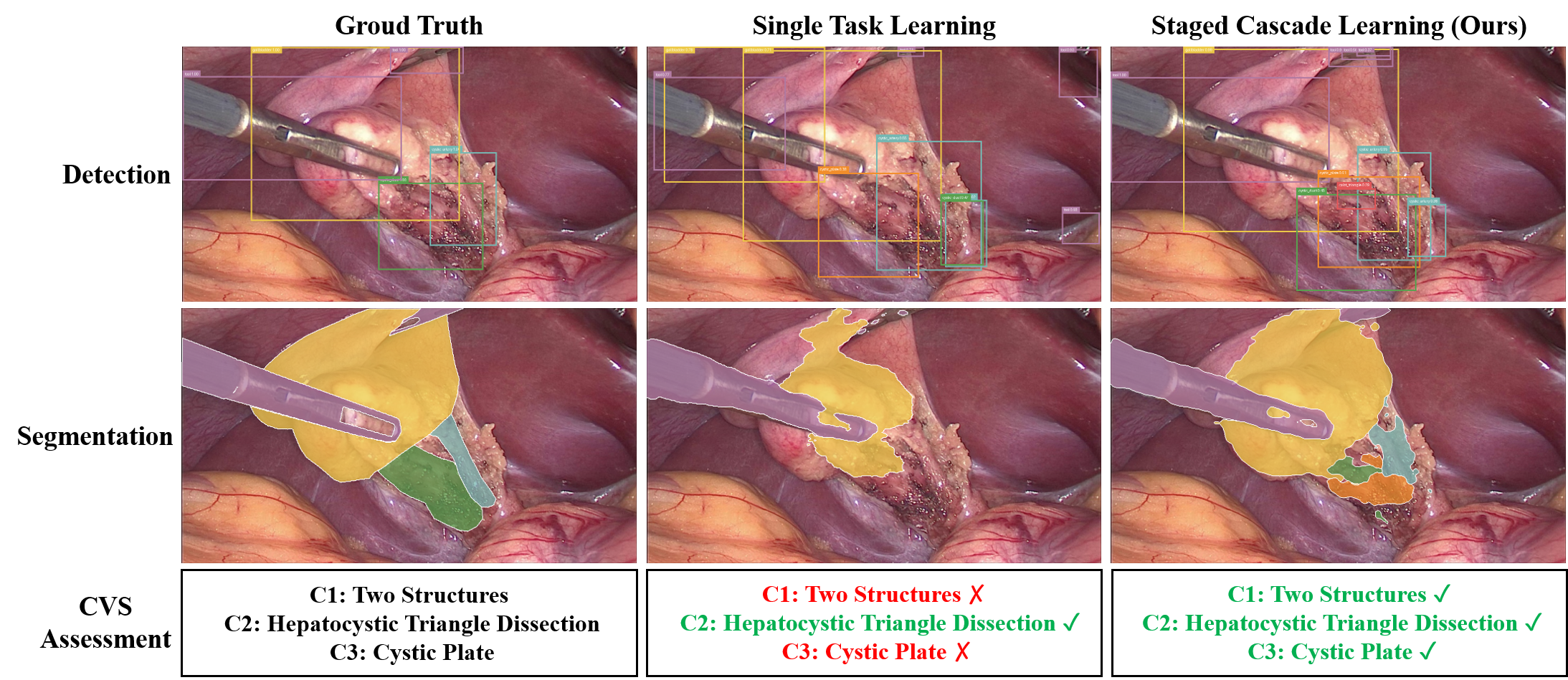}
    \caption{Single-task learning (STL) versus Staged Cascade Learning (Ours) on a test frame; columns: ground truth, STL, Ours; rows: detection and segmentation, with the~three CVS criteria (C1 Two Structures; C2 Hepatocystic Triangle Dissection; C3 Cystic~Plate) below each column. Each criterion is marked \textcolor{green!50!black}{green~\checkmark} when the prediction is correct and \textcolor{red}{red~\ding{55}} when it is wrong, with the reference labels shown under ground truth. STL misses the rare safety-critical anatomy and mispredicts the criteria, whereas Ours recovers the anatomy and matches the reference.}
    \label{fig:teaser}
\end{figure}

\noindent\textbf{Comparison with single-task and published methods.} Table~\ref{tab:main} reports test performance. Relative to single-task learning, the staged cascade improves every head: detection rises from 29.6 to 32.0, segmentation of the rare anatomy from 4.6 to 15.3 mIoU, and CVS from 47.9 to 67.2 mAP. The detection branch alone reaches 32.0 mAP, competitive with the strongest published detector on Endoscapes, Deformable-DETR (32.7), and above Faster R-CNN (30.0)~\cite{murali2023endoscapes}. The CVS mAP of 67.2 exceeds both the single-frame LG-CVS (63.2) and the spatiotemporal SV2LSTG (64.3) under the same public annotation setting, and the balanced accuracy of 74.7 is on par with LG-CVS (74.8) and above SV2LSTG (73.4). A single-frame model thus matches or surpasses a spatiotemporal one on this benchmark. Figure~\ref{fig:teaser} shows a qualitative comparison on a test frame.

\smallskip
\noindent\refstepcounter{table}\label{tab:main}\textbf{Table~\thetable.} Test performance on Endoscapes (\%). DET and SEG are detection mAP@[0.5:0.95] and semantic mIoU; rDET and rSEG are their rare-class values (cystic plate and hepatocystic triangle); CVS is reported as mAP and balanced accuracy (bAcc). Dashes denote metrics a method does not report; the segmentation mIoU is comparable only between the single-task model and ours. $^*$ spatiotemporal.\par
\begin{center}
\small
\setlength{\tabcolsep}{5pt}
\begin{tabular}{lcccccc}
\hline
 & & & & & \multicolumn{2}{c}{CVS} \\
\cline{6-7}
Method & DET & rDET & SEG & rSEG & mAP & bAcc \\
\hline
LG-CVS~\cite{murali2023lgcvs} & -- & -- & -- & -- & 63.2 & 74.8 \\
SV2LSTG$^*$~\cite{murali2023sv2lstg} & -- & -- & -- & -- & 64.3 & 73.4 \\
Single-task (STL) & 29.6 & 13.4 & 38.4 & 4.6 & 47.9 & -- \\
\textbf{Ours} & \textbf{32.0} & \textbf{16.4} & 46.8 & \textbf{15.3} & \textbf{67.2} & 74.7 \\
\hline
\end{tabular}
\end{center}

\noindent\textbf{Staging ablation.} Table~\ref{tab:staging} compares the staged schedule against single-stage joint training of the same network, with identical couplings and PCGrad, from scratch. Without staging, detection fails to train (3.9 mAP), leaving the downstream tasks with unreliable localisation cues; rare-anatomy segmentation falls to 0.1 mIoU and CVS to 28.0 mAP. Training detection first is therefore essential for stable cascade training.

\smallskip
\noindent\refstepcounter{table}\label{tab:staging}\textbf{Table~\thetable.} Staging ablation (test, \%): the same network and couplings trained jointly from scratch, without the staged schedule.\par
\begin{center}
\small
\setlength{\tabcolsep}{6pt}
\begin{tabular}{lcccc}
\hline
Training & DET & SEG & rSEG & CVS \\
\hline
Staged cascade (Ours) & \textbf{32.0} & \textbf{46.8} & \textbf{15.3} & \textbf{67.2} \\
Single-stage joint & 3.9 & 27.2 & 0.1 & 28.0 \\
\hline
\end{tabular}
\end{center}

\noindent\textbf{Cascade ablation.} Table~\ref{tab:ablation} ablates each coupling in turn while keeping the rest of the training recipe fixed. Removing the box-to-mask prior reduces rare-anatomy segmentation from 15.3 to 14.4 mIoU and CVS from 67.2 to 62.1 mAP. This suggests that box guidance helps recover the small hepatocystic structures on which CVS assessment depends, although the full model has slightly lower aggregate segmentation mIoU than this ablated variant (46.8 vs 48.0). Removing the mask-to-CVS coupling is far more damaging to CVS (67.2 to 55.8 mAP), even though this variant attains the \emph{highest} rare-anatomy mIoU (15.7). Thus, better segmentation alone does not improve CVS unless the classifier is explicitly coupled to the predicted anatomy. The mask-to-CVS coupling is therefore the main contributor to CVS performance, while the two couplings remain complementary rather than redundant. Overall, these couplings improve the downstream tasks only when the detector localises reliably, and cannot compensate for one that has not converged (Table~\ref{tab:staging}). Rare anatomy therefore remains the main segmentation bottleneck even in the full model (15.3 vs 46.8 mIoU).

\smallskip
\noindent\refstepcounter{table}\label{tab:ablation}\textbf{Table~\thetable.} Cascade ablation (test, \%), removing one coupling at a time; metrics as in Table~\ref{tab:main}.\par
\begin{center}
\small
\setlength{\tabcolsep}{6pt}
\begin{tabular}{lcccc}
\hline
Configuration & DET & SEG & rSEG & CVS \\
\hline
Full (both couplings) & 32.0 & 46.8 & 15.3 & \textbf{67.2} \\
\quad without box-to-mask & 32.2 & 48.0 & 14.4 & 62.1 \\
\quad without mask-to-CVS & 29.6 & 48.4 & \textbf{15.7} & 55.8 \\
\hline
\end{tabular}
\end{center}

\section{Conclusion}
\label{sec:conclusion}

CasCVS-Net combines the complementary tasks of detection, segmentation, and CVS assessment: detection localises anatomy, segmentation produces masks, and a mask-pooled head assesses the three criteria. The staged cascade improves all tasks over single-task baselines, with the largest gain in rare-anatomy segmentation (4.6 to 15.3 mIoU). Its 67.2 CVS mAP exceeds single-frame LG-CVS and spatiotemporal SV2LSTG by 4.0 and 2.9 points, respectively (6.3\% and 4.5\% relative); ablations identify mask-to-CVS coupling as the main contributor. The method requires a converged detector and processes frames independently; rare anatomy remains hardest to segment. Next steps are temporal modelling and multi-centre validation.

\end{document}